\documentclass{article}
\usepackage{ijcai25}

\usepackage{times}
\usepackage{soul}
\usepackage{url}
\usepackage[hidelinks]{hyperref}
\usepackage[utf8]{inputenc}
\usepackage[small]{caption}
\usepackage{graphicx}
\usepackage{amsmath}
\usepackage{amsthm}
\usepackage{amsfonts}
\usepackage{booktabs}
\usepackage{algorithm}
\usepackage{algorithmic}
\usepackage[switch]{lineno}
\usepackage{comment}
\usepackage{dirtytalk}
\usepackage{dsfont}
\usepackage{natbib}

\title{Neural Spatiotemporal Point Processes: Trends and Challenges}

\author{
Sumantrak~Mukherjee$^1$ \and
Mouad~Elhamdi$^{1,4}$\and
George~Mohler$^{2}$\and
David~A.~Selby$^{1}$\and \\
Yao Xie$^{5}$\and
Sebastian~Vollmer$^{1,3}$ 
\And 
Gerrit~Großmann$^1$\\
\affiliations
$^1$German Research Center for Artificial Intelligencex\\
$^2$Boston College\\
$^3$University of Kaiserslautern--Landau\\
$^4$Mohammed VI Polytechnic University \\
$^5$Georgia Institute of Technology
\emails
\{sumantrak.mukherjee, david\textunderscore antony.selby, sebastian.vollmer, gerrit.grossmann\}@dfki.de,
mouad.elhamdi@um6p.ma,
mohlerg@bc.edu, 
yao.xie@isye.gatech.edu
}

\begin{document}

\maketitle

\begin{abstract} 
Spatiotemporal point processes (STPPs) are probabilistic models for events occurring in continuous space and time. Real-world event data often exhibit intricate dependencies and heterogeneous dynamics. By incorporating modern deep learning techniques, STPPs can model these complexities more effectively than traditional approaches. Consequently, the fusion of neural methods with STPPs has become an active and rapidly evolving research area. In this review, we categorize existing approaches, unify key design choices, and explain the challenges of working with this data modality. We further highlight emerging trends and diverse application domains. Finally, we identify open challenges and gaps in the literature.
 \end{abstract}

\section{Introduction}
\label{sec:introduction}
Real-world events---such as urban crime incidents, epidemic spread, earthquakes, and environmental changes---can be represented as sequences of discrete events with both spatial and temporal components.
Studying the spatiotemporal distribution of events and discovering the relationships among different types of events is an increasingly important area of research for understanding the dynamics and mechanism of the occurrence of the events. One such paradigm is the \emph{spatiotemporal point process} (STPP) model, defined as a stochastic process that describes the spatial and temporal distribution of discrete events \citep{daley2007introduction}, which is well suited to capture the complex relationships between events, including self-excitation, and the interactions between events and spatial covariates across time and space.

Reviews on neural point processes have focused on modeling the temporal dynamics of events using neural networks (NNs) \citep{shchur2021neural,lin2021empirical,lin2022exploring}. In contrast, reviews that address spatiotemporal event modeling have focused on traditional statistical methods \citep{gonzalez2016spatio,reinhart2018review}. More recently, some reviews have explored specific aspects of the use of machine learning in STPPs \citep{wikle2023statistical,bernabeu2024spatio}, these typically focus on specific aspects rather than providing comprehensive coverage. Meanwhile, work on neural STPPs has achieved remarkable progress. 

Our work bridges this gap by systematically exploring design choices, methodological innovations, and key challenges in neural STPPs. To our knowledge, no prior survey has comprehensively examined these aspects in this context.

\begin{figure}[t]
    \centering
    % Use \linewidth (or \columnwidth) to make the figure occupy 100% of the column's width
    \includegraphics[width=0.8 \linewidth]{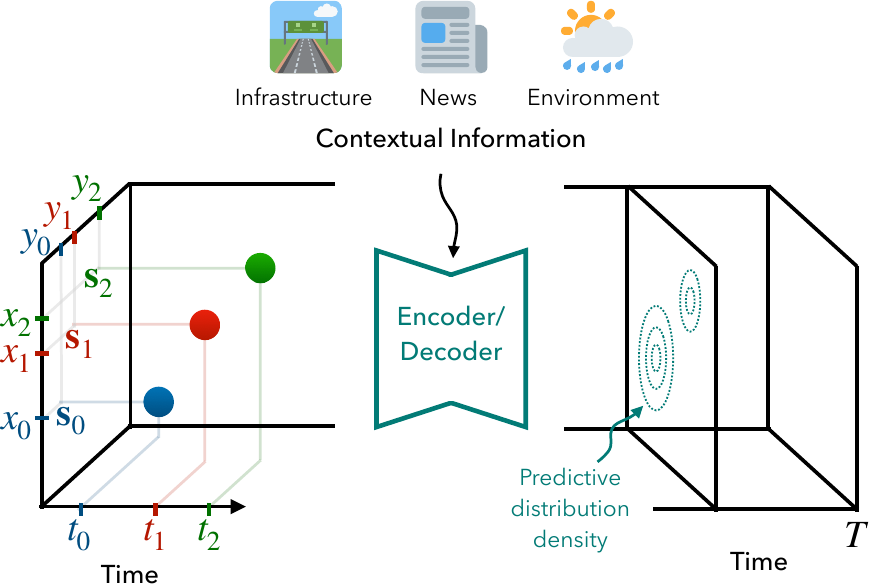}
    \caption{Schematic of the autoregressive construction of an STPP with two spatial dimensions. Three events are fed into an NN that predicts the likelihood of future event times and locations.}
    \label{fig:stppdef}
\end{figure}

\begin{figure*}[htbp]
    \centering
    \includegraphics[width=0.8\textwidth]{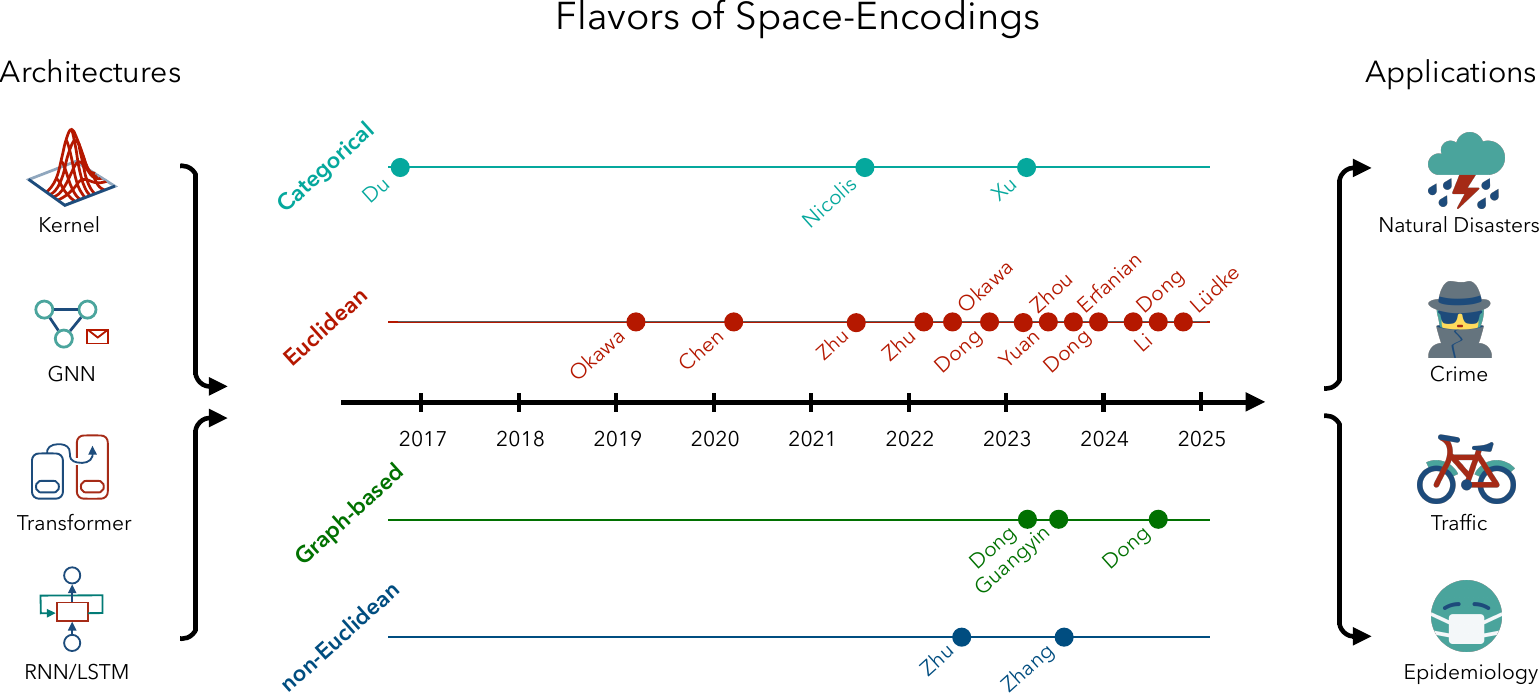}
    \caption{A timeline of the reviewed methodological and application-focused works, along with an overview of neural architectures and application domains. Papers are grouped based on how they encode spatial information, though in some cases these categorizations are not strictly defined.}
    \label{fig:main}
\end{figure*}

\paragraph{Why we need neural STPPs.}
Structural differences between space and time make modeling challenging. Time is unidirectional, while spatial propagation is omnidirectional and is affected by environmental factors. Traditional methods rely on strong parametric assumptions and independence, limiting flexibility. They struggle with long-range dependencies, fail to capture heterogeneous dynamics across space and time, and cannot integrate multimodal data.  
Additionally, single-step predictions cause error accumulation. NN-based methods overcome these limitations by encoding space and time efficiently, handling dependencies, and learning heterogeneous patterns. They automate feature extraction, integrate diverse data, and scale effectively. 

% \paragraph{Scope and structure of the paper.}
% This survey reviews neural STPPs, covering core models, applications, and key components in event modeling. We focus on studies that use point processes with neural parameterization to capture spatiotemporal dynamics. Our literature search included keyword-based queries, citation tracking, and seminal works in neural temporal point processes. By outlining fundamental principles and design choices, we provide a practical foundation for researchers.
% We first introduce the necessary background and notation and then present the available modeling choices, including architectures, training procedures and metrics. 
% The next section highlights the most notable applications of neural STPPs and we conclude with a section on open challenges.
\paragraph{Scope and structure of the paper.}
This survey reviews neural STPPs, covering core models, applications, and key components in event modeling. We focus on studies using point processes with neural parameterization to capture spatiotemporal dynamics. Our literature search included keyword-based queries, citation tracking, and seminal works in neural temporal point processes. By outlining fundamental principles and design choices, we provide a practical foundation for researchers.
We first introduce necessary background and notation, then present available modeling choices, including architectures, training procedures, and metrics.
The next section highlights notable applications of neural STPPs, and we conclude 
by discussing open challenges.

\section{Background and Notation}
\label{sec:Background and Notation}

\paragraph{Spatiotemporal point processes.} 
STPPs are concerned with modeling sequences of random events in continuous space and time \citep{moller2003statistical}. A realization or sample of an STPP is defined up to a \emph{time horizon} $T \in \mathbb{R}_{\geq 0}$. It is a finite, ordered \emph{event sequence} containing pairs \( X = [(t_1, \boldsymbol{s}_1), (t_2, \boldsymbol{s}_2), \dots, (t_n, \boldsymbol{s}_n)] \), where $t_i \in [0,T]$ denotes the time and \( \boldsymbol{s}_i \in \mathcal{S} \) the location of event $i$. Here, \( \mathcal{S} \subseteq \mathbb{R}^d \) represents the spatial domain, which is typically a bounded region in \( d \)-dimensional Euclidean space and typically $d=2$. For a given event sequence, $\mathcal{H}_t = \lbrace (t_i, \boldsymbol{s}_i) \mid t_i < t \rbrace$ denotes the history of the current realization up to (but excluding) time point $t$ (c.f.\ Figure~\ref{fig:stppdef}).

\paragraph{Likelihood.}  
An STPP can be specified by defining the likelihood of event sequences. The framework follows the \emph{temporal priority principle} \citep{berzuini2012causality}, which asserts that all causes must precede their effects. As a result, the likelihood of an event sequence \( X \) can be expressed in an auto-regressive form \citep{rasmussen2018lecture}:
\[
f(X) = \underbrace{\left( \prod_{i=1}^{n} f^{\text{pred}}(t_i, \boldsymbol{s}_i \mid \mathcal{H}_{t_i}) \right)}_{\text{Likelihood of observed events}} \cdot \underbrace{\left(1 - F^{\text{pred}}(T \mid \mathcal{H}_{t_n}) \right)}_{\text{Probability of no events after } t_n},
\]
where \( f^{\text{pred}}(t, \boldsymbol{s} \mid \mathcal{H}_t) \) is the \emph{predictive distribution}, specifying the conditional probability density function (PDF) for the next event occurring at a given timestamp \( t \) and location \( \boldsymbol{s} \), given the history of past events \( \mathcal{H}_t \). The term \( F^{\text{pred}}(T \mid \mathcal{H}_{t_n}) \) represents the cumulative distribution function (CDF) of the predictive distribution, which gives the probability that an event occurs before or at time \( T \), regardless of \( \boldsymbol{s} \).
%, obtained by integrating the PDF over all possible locations and up to time \( T \).
Thus, \( 1 - F^{\text{pred}}(T \mid \mathcal{H}_{t_n}) \) is the probability of no events occurring after the last observed event.  %.. at time \( t_n \).
%, regardless of the spatial location.
Conveniently, this formulation also provides a principled approach for simulation (i.e., the generation of samples from the underlying stochastic model).

\paragraph{Intensity function.}  
In practice, an STPP is often described using a \emph{(conditional) intensity function} (CIF) instead of a predictive distribution; they are mutually translatable \citep{chen2021neural}. 
The CIF \(\lambda(t, \boldsymbol{s}\mid\mathcal{H}_t)\), denoted by \(\lambda^{*}(t, \boldsymbol{s})\) as a shorthand for its dependence on the\(\mathcal{H}_t\), is defined as:

{\small
\begin{equation}
\begin{split}
 \lambda^{*}(t, \boldsymbol{s}) = \lim_{\Delta_{\boldsymbol{s}}, \Delta_t \downarrow 0} 
    \frac{\mathbb{P} \big( \text{Event} \in \big( B(\boldsymbol{s}, \Delta_{\boldsymbol{s}}) \times [t, t + \Delta_t) \big)
    \mid \mathcal{H}_{t} \big)}{ \left| B(\boldsymbol{s}, \Delta_{\boldsymbol{s}}) \right| \Delta t},
\end{split}
\label{eq:conditional_intensity}
\end{equation}}

where \( B(\boldsymbol{s}, \Delta_s) \) is a \( d \)-dimensional ball (a disk if \( d=2 \)) centered at \( \boldsymbol{s} \in \mathcal{S} \) with radius \( \Delta_s\), and \( |B(\boldsymbol{s}, \Delta_s)| \) denotes its volume.

Events can include additional information beyond location as a \emph{mark} \citep{daley2007introduction}. While the core concept remains unchanged, this approach incorporates multimodal contextual data

\section{Modeling Neural Spatiotemporal Point Processes}
\label{sec:Modeling Neural Spatiotemporal Point Processes}
Neural STPPs model event evolution and probabilities using NNs, implicitly encoding a PDF over sequences. The chosen approach affects training efficiency, likelihood evaluation, and event generation. A common method constructs latent representations from an event history to generate new events. Adressing the challenges of spatial encoding, we first review methods designed to address these challenges, then discuss multi-event generation.
%We first review these methods and then discuss multi-event generation and spatial encoding challenges.
We do not focus on neural methods for temporal encoding as they are well-studied \cite{shchur2021neural}.
%, as temporal encoding is well-studied.

Modeling choices depend on application needs. Crime models may focus on landmark influence, while earthquake prediction prioritizes accuracy. Whereas, epidemic models may emphasize county-level policy, reducing spatial granularity. Some applications require parametric assumptions for hypothesis testing, while safety-critical settings demand uncertainty quantification. These considerations shape spatial encoding strategies. While discretization simplifies modeling, it reduces the ability to encode an inductive bias where nearby events exert stronger influence. Its performance is also sensitive to granularity, boundary effects, and local correlations.
Graph-based methods like GNNs mitigate these issues by modeling inter-cell influences.

For continuous-space methods, using raw coordinates (e.g., latitude-longitude) ignores anisotropic propagation. Solutions include location-specific parametric kernels, contextual data (e.g., satellite images, geotagged text), and learned spatial embeddings. Attention models implicitly capture heterogeneous effects, while GNNs and heterogeneous kernels explicitly model spatial interactions. 

\subsection{History Event Encoder}
\label{sec:historyencoder}
Spatiotemporal event modeling requires a CIF that combines temporal evolution and spatial dependencies (c.f.\ Figure \ref{fig:stppdef}). 
A common approach to predicting an event \((t_i,\boldsymbol{s}_i)\) is to first encode each historical event 
\((t_j,\boldsymbol{s}_j) \in \mathcal{H}_{t_i}\) as an embedding 
\(
\boldsymbol{e}_j = [\omega(t_j);\sigma(\boldsymbol{s}_j)].
\)
Some architectures directly use raw time \( t_j \) and space \( \boldsymbol{s}_j \), while others transform them via \( \omega(\cdot) \) (e.g., linear, trigonometric, or logarithmic mappings) and \( \sigma(\cdot) \) (e.g., linear layers or one-hot encodings). A history encoder then processes embeddings into a latent state \( \boldsymbol{h}_i \) to parameterize the CIF .

RNN variants like GRU and LSTM are common history encoders, updating states as \( \boldsymbol{h}_{i+1} = \text{RNN}(e_i, \boldsymbol{h}_i) \) \citep{du2016recurrent,omi2019fully,Shchur2020Intensity-Free}. Their sequential nature reduces storage needs but limits parallelization. Additionally, they suffer from gradient vanishing and long-term memory loss \citep{le2016quantifying}.

Attention mechanisms \citep{vaswani2017attention} overcome several limitations of recurrent encoders. They have demonstrated superior performance as history encoders for temporal point processes \citep{zhang2020self,zuo2020transformer} and lately in STPPs \citep{zhou2022neural,yuan2023spatio}. Nevertheless, the $\mathcal{O}(N^2)$ space complexity required to construct the attention matrix can pose practical challenges.% when dealing with very long event sequences.

\subsection{Single Event Prediction}

 Neural STPPs predict the time and location of the next event. We review parametric, neural, mixture, diffusion, and continuous-time models, emphasizing those that capture spatial heterogeneity, contextual influences, and graph-based interactions.

\paragraph{Kernel-based methods.}
Kernel-based methods model spatiotemporal dependencies by parameterizing event influence through kernels. When combined with NNs, these kernels become more flexible while maintaining interpretability. The intensity function is typically defined as follows:
\begin{equation}
\lambda^*(t, \boldsymbol{s}) = \mu(t, \boldsymbol s) + \sum_{(t', \boldsymbol{s}') \in \mathcal{H}_t} K(t', t, \boldsymbol{s}', \boldsymbol{s}),
\label{eq:Generalized_Kernel}
\end{equation}
where $\mu (t, \boldsymbol s)$ is the baseline rate (which can potentially vary over time and space), and $K(\cdot)$ models past event influence. Traditional models use stationary kernels, assuming time and space invariant relationships (e.g., ETAS \citep{musmeci1992space}, \citep{hawkes1971spectra}), making strong parametric assumptions and missing heterogeneous effects. More expressive approaches learn spatial and temporal dependencies jointly or independently, improving adaptability.
  
\paragraph{Non-stationary kernels.}  
Designing the influence kernel $K(\cdot)$ in Equation~\eqref{eq:Generalized_Kernel} is crucial for capturing how past events trigger future occurrences. Neural parameterizations of $K(\cdot)$ enable non-stationary dependencies that vary across space, time, and contextual marks, moving beyond fixed parametric forms.  

A representative approach is the Gaussian mixture model of \citet{zhu2021imitation}, where $K(\cdot)$ is expressed as:  
\begin{equation}
    K(t', t, \boldsymbol{s}',\boldsymbol{s}) = \sum_{l=1}^{L} \phi_{\boldsymbol{s}'}^{(l)} \cdot g(t,t',\boldsymbol{s},\boldsymbol{s}' \mid \Sigma_{\boldsymbol{s}'}^{(l)}, \mu_{\boldsymbol{s}'}^{(l)}).
    \label{eq:NESTkernel}
\end{equation}  
A neural network embeds spatial coordinates $\boldsymbol{s}$ to generate location-specific parameters $\mu_{\boldsymbol{s}}^{(l)}$, $\Sigma_{\boldsymbol{s}}^{(l)}$, and mixture weights $\phi_{\boldsymbol{s}}^{(l)}$ where $l \in 1,\dots,L$ with $L$ denoting number of mixture components. Applying constraints ensures physical interpretability and reflects heterogeneous event diffusion. Visualizing learned kernels reveals region-specific influence propagation, making $K(\cdot)$ a smooth, adaptive function.

Alternate formulations consider heterogeneous interactions between events modeled using Mercer's theorem and neural basis functions\citep{zhu2022neural}, leverage low-rank decomposition and a deep non-stationary influence kernel \citep{dong2022spatio}, or embed events in graphs for non-Euclidean interactions \citep{dong2023deep}. These can be generalized as:  
\begin{equation} 
K(t',t,\boldsymbol{s}',\boldsymbol{s}) = 
\sum_{r=1}^R \sum_{l=1}^L \alpha_{rl}\psi_l(t',t)\phi_r(\boldsymbol{s}',\boldsymbol{s}), 
\label{eq:GeneralKernel} 
\end{equation}  
where $\psi_l(\cdot)$ and $\phi_r(\cdot)$ are neural feature maps modeling time displacement, spatial relations, or graph connectivity, with $\alpha_{rl}$ scaling their contributions. $L$ and $R$ refer to the number of basis kernels used to decompose the kernel in Equation \eqref{eq:Generalized_Kernel}. 

For instance, \citet{dong2023non} and \citet{dong2024atlanta} use learned Gaussian bases for anisotropic geography, while \citet{dong2024spatio} combine a stationary Gaussian kernel with a Graph Neural Network (GNN) mark kernel to capture network constraints and landmark effects.

\paragraph{Context.} Event dynamics are influenced by contextual covariates like georeferenced images and text. \citet{okawa2019deep} define event intensity as a spatially localized mixture of kernels:
\begin{equation}
\lambda^*(t, \boldsymbol{s}\mid \mathcal{D}) = \sum_{j=1}^{M} f(\boldsymbol{u}_j, \boldsymbol{z}_j; \theta) K(t, \boldsymbol{s}, \boldsymbol{u}_j),
\label{eq:DMPPintensity}
\end{equation}
where $K(\cdot)$ is a compactly supported Gaussian kernel, $\mathcal{D}$ is a set of contextual features, and $\boldsymbol{u}_j$ are spatiotemporal anchors uniformly distributed in time and space. Contextual features $\boldsymbol{z}_j$, extracted from $\boldsymbol{u}_j$, inform mixture weights $f(\boldsymbol{u}_j, \boldsymbol{z}_j; \theta)$ through a deep network combining image and text embeddings. These weights adapt to heterogeneous conditions (e.g., urban infrastructure, social events), enabling dynamic kernel weighting centered on $u_j$ while restricting contextual influence to local neighborhoods.

\citet{zhang2023integration} replace the parametric kernel with a deep kernel $K(t, \boldsymbol{s},\boldsymbol{u}) = k_\phi(g(t, \boldsymbol{s}), g(\boldsymbol{u}))$, where $g(\cdot)$ is a non-linear transformation by a deep NN, learning complex spatial correlations beyond standard Euclidean distance.

\citet{okawa2022context} propose a method for incorporating high-dimensional contextual data into the Hawkes process. They introduce a weighting term in the excitation kernel, extracting relevant features using CNNs and employing continuous kernel convolution to transform discretized features into continuous space. This enables capturing spatial heterogeneity and external influences while ensuring tractable optimization.

\paragraph{Semi-parametric and non-parametric kernels.} 
\citet{zhou2022neural} introduce a non-parametric mixture-based intensity given by
\begin{equation}
    \lambda^*(\boldsymbol{s},t \mid \boldsymbol{z}) \;=\; \sum_{i=1}^{n+J} w_i \, K_{\boldsymbol{s}}\bigl(\boldsymbol{s},\boldsymbol{s}_i;\gamma_i\bigr)\,K_t\bigl(t,t_i;\beta_i\bigr),
    \label{eq:DeepSTPP}
\end{equation}
where each kernel $K_{\boldsymbol{s}}(\cdot)$ and $K_t(\cdot)$ is a normalized radial basis function, and the parameters $\{w_i, \gamma_i, \beta_i\}$ are drawn from a latent process $\boldsymbol{z}$. The history is encoded by a Transformer \ref{sec:historyencoder} and decoded to parameterise the latent process. The samples of the latent process are further decoded via a feedforward network. This method captures uncertainty in the timing and spatial locations of events. By augmenting the observed $n$ events with $J$ randomly sampled representative points, this approach addresses global background intensity, thereby reducing the reliance on strong parametric assumptions.

Neural STPP models often rely on restrictive assumptions, such as conditional independence or unilateral dependence between the distributions of temporal and spatial events. These assumptions limit their ability to accurately predict events in real-world scenarios, where events exhibit complex interdependencies in both time and space. To address this, \citet{yuan2023spatio} proposed a framework that jointly models spatiotemporal event distributions via a diffusion-based approach without structural constraints. The model employs a spatiotemporal encoder that separately embeds time and space, fuses these into spatiotemporal representations using self-attention, and conditions a diffusion model on these hidden states. The diffusion process iteratively denoises event coordinates using a co-attention network that dynamically captures cross-dependent spatiotemporal interactions, enabling joint distribution learning without assuming independence or requiring integrable intensity functions. For spatial decoding, the model directly predicts continuous coordinates or can apply a rounding step for discrete locations. This method eliminates the need for approximation during sampling and supports continuous and discrete spatial domains.

Despite the inherent randomness in event times and locations, many STPP models offer only point predictions, lacking principled uncertainty quantification. This gap is especially problematic in marked STPPs, where reliable confidence scores for discrete event marks are essential. To address these challenges, \citet{li2024beyond} 
introduce a score-matching objective for estimating the pseudo-likelihood of marked STPPs overcoming issues with intractable integral calculations while also providing uncertainty estimates
the score function represents the gradient of the logarithm of the conditional spatial distribution. 
They use the same CDN architecture as the backbone as \cite{yuan2023spatio} to predict the score function.
Langevin dynamics is employed to sample event locations by iteratively refining draws according to the learned score. Thresholding the resulting sample density yields confidence regions for event locations, while a similar procedure provides confidence intervals for event times. The proposed framework not only predicts future events accurately, but also quantifies uncertainty, offering robust confidence bounds for both the event timing and discrete marks.

\paragraph{Continuous time-based methods.} The work by \citet{chen2021neural} employs Continuous-time Normalizing Flows (CNF), which is based on Neural Ordinary Differential Equations \citep{chen2018neural}. This method separates the conditional CIF into two components temporal component and a spatial conditional distribution. Subsequent developments of CNFs further refine the spatial conditional distribution. For instance, the Time-Varying CNF introduces continuous flow transformations to handle time-varying observational data; however, it does not explicitly incorporate conditioning on the event history.

To address this limitation, Jump CNFs combine insights from \citet{jia2019neural} with CNFs, integrating discrete state updates at event times. This efficiently models abrupt dynamic changes with a computational cost of 
$\mathcal{O}(N)$ for
$N$ past events. To improve scalability for long event histories, Attentive CNFs use Transformer-based attention, enabling parallel trajectory computation while preserving non-trivial dependencies, achieving a balance between efficiency and representational power.

\subsection{Multi-Event Prediction}

Single-event prediction methods rely on stepwise CIF estimation, making multi-event forecasting computationally prohibitive in high-dimensional spaces due to repeated integration. Sequential history updates also propagate errors, leading to degraded accuracy over iterations. Multi-event prediction addresses these issues by jointly estimating event distributions, eliminating reliance on sequential updates. Transformer-based architectures enable parallelized predictions, bypassing error accumulation and improving efficiency.

\paragraph{Multi-step decoders.}
\citet{erfanian2022beyond} propose a new Transformer-based architecture augmented with normalizing flows and probabilistic layers, which outputs a batch of $L$ events based on the history of previous $n$ events.
In this approach, a separate spatial and temporal encoding is learned for events ranging from $n+1$ to $n+L$, which are then injected into a probabilistic layer through a learned mapping between the parameters of an exponential distribution for time and a multivariate Gaussian distribution for space.
Considering these as the base distributions they are passed into normalizing flows, which convert them into more expressive distributions to model the joint distribution of batched events.
The parameters of all distributions are learned independently for $l \in [n+1, \dots, n+L]$.
Even though separate distributions are used for space and time, the hidden state used for both as inputs considers spatiotemporal interdependencies. 

\paragraph{Sampling full sequences.}
Another approach models the probability of the entire sequence or point set, rather than modeling the inter-event time and spatial distribution given history. This approach addresses a key problem with CIF parameterization methods. In this modeling framework, the entire event sequence is embedded in the analysis. Moving away from the CIF parameterization offers added benefits for modeling events, such as data imputation and multi-event prediction.
\citet{ludke2023add} model entire TPPs using diffusion models. Superposition and thinning properties of the CIF define the noising (forward) and denoising (backward) methods. In a recent study, \citet{ludke2025unlocking} extends this work to point processes defined on general metric spaces while generalizing to order space, like STPPs.  This work further develops the method, enabling flexible conditioning on various tasks without the need for explicit task-specific training. The approach separates the thinning and superposition operations into two independent processes.
%, providing greater control over the model dynamics. 
This separation allows the diffusion process to be defined as a stochastic interpolation between two point sets, entirely independent of the intensity function.

\begin{table*}[tp]
\centering
\small
\caption{Common Evaluation Metrics for Neural Spatiotemporal Point Processes}
\label{tab:evaluation_metrics}
\begin{tabular}{p{0.13\textwidth}p{0.38\textwidth}p{0.38\textwidth}}
\toprule
\textbf{Name} & \textbf{Definition} & \textbf{Advantages} \\ 
\midrule

\textbf{NLL} 
& Negative Log-Likelihood (likelihood of observed data given the predicted distribution). 
& Directly tied to MLE-based training; widely adopted for distributional fit. \\[0.3em]

\textbf{HD} 
& Hellinger Distance (a measure of distance between two probability distributions). 
& Fine-grained assessment of similarity; often used with known ground-truth distributions. \\[0.3em]

\textbf{MMD} 
& Maximum Mean Discrepancy (kernel-based comparison of two sample sets). 
& Distribution-agnostic; captures higher-order differences in generative quality. \\[0.3em]

\textbf{MAE / MSE / RMSE} 
& Mean/Absolute/Squared/Root Errors for point predictions (time or space). 
& Straightforward and interpretable; highlight large errors (MSE/RMSE) or typical errors (MAE). \\[0.3em]

\textbf{MAPE} 
& Mean Absolute Percentage Error (ratio-based prediction error). 
& Scale-independent; intuitive for relative errors across different magnitudes. \\[0.3em]
\textbf{SL} 
& Sequence Length (Wasserstien Distance between the categorical distribution of event sequence length ). 
& Suitable for applications where event count within an interval outweighs spatial and temporal accuracy. \\[0.3em]

\textbf{CD} 
& Counting Distance (A generalisation of Wasserstien Distance for order TPPs to STPP using L1 distance)
& Useful for evaluating the generative performance of model, especially for multi-step predictions \\[0.3em]

\textbf{CS / ECE} 
& Calibration Score / Expected Calibration Error (comparison of predicted vs.\ observed confidence). 
& Evaluates how well probability estimates reflect true event frequencies (calibration). \\

\bottomrule
\end{tabular}
\end{table*}

\section{Parameter Estimation and Inference}
\label{sec:Parameter Estimation and Inference}

The key objectives of event prediction include enhancing predictive performance, improving generalization and robustness, understanding event dynamics through learned parameters, accounting for event behavior heterogeneity, and capturing the influence of external factors. Predicting future events from a learned model requires sampling from its intensity function. Traditional statistical methods often rely on strong parametric assumptions for modeling event intensities, using techniques such as likelihood-based methods, partial likelihood, the EM algorithm, or Bayesian approaches.

In statistical inference, \textbf{Maximum Likelihood Estimation} (MLE) is most commonly used to fit classical and neural STPPs, typically by maximizing the likelihood function or, equivalently, minimizing the negative log-likelihood (NLL). For an observed sequence of $N$ events, the NLL is given by \citep{daley2003introduction}:

{\small
\[
\mathcal{L}_{NLL} = -\underbrace{\sum\limits_{i=1}^{N} \log \lambda^{*}(t_i, \boldsymbol{s}_i)}_{\text{Log-likelihood of observed events}} + \underbrace{\int\limits_{[0,T]} \int\limits_{\mathcal{S}} \lambda^{*}(\tau, \boldsymbol{u})\, d\tau\, d\boldsymbol{u} \, .}_{\text{Expected number of events}}
\]
}

When using neural networks to parameterize the CIF, evaluating the integral term is typically intractable. This complexity often requires numerical methods \citep{chen2021neural} or Monte Carlo methods \citep{mei2017neural} for likelihood evaluation. However, these strategies can be computationally expensive and prone to numerical errors, particularly in high-dimensional spatiotemporal domains. While certain simplifying assumptions, such as exponential decay \citep{du2016recurrent} and linear interpolation \citep{zuo2020transformer}, can lead to closed-form solutions or faster approximations, they often restrict the expressiveness of the model.

In the purely temporal settings, \citet{zhou2023automatic} introduce a paradigm for efficient and non-parametric inference of TPPs. They approximate the influence function (c.f.\ Equation \eqref{eq:Generalized_Kernel}) via a NN, using \textbf{automatic integration} to compute its integral. A monotonically increasing integral network is trained, its partial derivative defines the CIF. This approach directly yields the CIF and its antiderivative from the network parameters, avoiding functional form restrictions. 
Building on this foundation, \citet{zhou2024automatic} addresses the computational challenge of integrating the intensity function in $3$D spatiotemporal domains by employing automatic integration. This approach learns an integral network, whose partial derivatives with respect to spatial and temporal inputs yield the intensity, ensuring an exact antiderivative without restricting the model’s functional form. Furthermore, a ProdNet factorization of the influence function into 1D components enforces non-negativity while capturing spatiotemporal interactions. Maximizing the log-likelihood of observed events learns a highly expressive CIF, improving spatiotemporal event prediction.

\citet{zhou2022neural} integrates flexible non-parametric modeling with \textbf{amortized variational inference}. This model captures events in continuous time, capturing irregular sampling dynamics and unifying spatial and temporal dependencies. By employing a kernel-based intensity function, the approach allows for closed-form integration, addressing previously intractable likelihood computations. This design avoids computationally expensive numerical integration inherent in neural ODE-based approaches \citep{chen2021neural}. Furthermore, this non-parametric approach avoids restrictive parametric assumptions. Training via amortized variational inference maximizes the evidence lower bound (ELBO) of the likelihood while balancing reconstruction accuracy and posterior regularization. The framework uses the kernel-based intensity for gradient computation, facilitating end-to-end optimization of both encoder and decoder parameters.

\citet{zhang2023integration} uses \textbf{score matching}, which minimizes the Fisher divergence between the model's log-density gradient and the data's log-density gradient, thus bypassing the need to calculate the intractable integral. A denoising score matching (DSM) method is used, which improves stability by introducing a small amount of noise to the data, which avoids the computation of second derivatives. In contrast, \citet{li2024beyond} utilizes a score matching-based pseudolikelihood objective, which eliminates the need for explicit calculation of the normalizing term that makes the likelihood integral intractable. The model decomposes the joint intensity function into a product of conditional distributions, allowing for the application of score-matching techniques for event times and locations, while using a conditional likelihood for event marks. This approach is designed to overcome overconfidence and underconfidence by learning a posterior distribution that matches the actual data distribution, which also requires score-based sampling with Langevin dynamics.
 
\textbf{Reinforcement learning} (RL) frameworks provide a training approach that does not rely on likelihood calculations. \citet{zhu2021imitation} employs an imitation learning framework to train their model. The learner policy is defined by a PDF associated with the CIF of the point process and parameterized by the model's parameters. The goal is for this learner policy to replicate the expert policy reflected in the training data. The training process maximizes the expected reward, determined by the Maximum Mean Discrepancy (MMD) between empirical distributions of the training data and data generated by the learner's policy. This data-driven MMD reward function offers robustness to model mismatch as it compares data distributions instead of relying on a predefined likelihood. Additionally, the closed-form representation of the reward function enables computationally efficient optimization via analytical gradient calculation, avoiding intensive inverse reinforcement learning.

\section{Evaluation Metrics}
\label{sec:Evaluation Metrics}
Evaluating STPPs requires task-specific metrics, summarized in Table \ref{tab:evaluation_metrics}. Model fit is often assessed using \textbf{NLL}, which measures how well predicted distributions align with observed data. However, NLL prioritizes overall distributional fit over individual event accuracy and can be biased due to computational approximations like Monte Carlo integration. Moreover, since NLL conditions on ground truth history, it is limited in evaluating true generative capacity.

Alternative distributional metrics address these shortcomings. \textbf{HD} \citep{zhou2022neural} directly compares learned and true distributions but requires ground truth intensity and spatial discretization, limiting real-world use but useful for testing the ability of models to recover known patterns. \textbf{MMD} \citep{zhu2021imitation} avoids strong distributional assumptions by comparing generated and observed sequences but is computationally expensive and sensitive to kernel selection.  MMD can also be used with \textbf{CD} \citep{ludke2025unlocking} as the distance measure. This improves distributional comparisons and is useful for evaluating the generative performance of STPPs. \textbf{SL}, measured via Wasserstein distance between two categorical distributions, provides additional insight into model performance by comparing event sequence lengths, especially useful when the task involves measuring case counts such as in epidemiology or crime modelling.

For point prediction accuracy, \textbf{MAE}, \textbf{MSE}, and \textbf{RMSE} are commonly used. MAE is robust to uniform errors, while RMSE is more sensitive to large deviations. Aggregated event predictions rely on normalized MAE (NMAE) \citep{okawa2022context} and \textbf{MAPE} \citep{okawa2019deep}. While NMAE enables cross-scale comparisons, it depends on predefined spatiotemporal regions. MAPE, though intuitive, can be unstable near zero values. Mean relative error (MRE) \citep{dong2022spatio} assesses intensity differences but also suffers from instability near zero. Prediction accuracy (ACC) is useful for event count estimation but, unlike metrics considering location and time, it assesses the accuracy of event counts only.
Prediction accuracy (ACC) is useful for event count estimation but doesn't account for spatial and temporal precision.

Uncertainty quantification is crucial for robust evaluation. \textbf{CS} and \textbf{ECE} \citep{li2024beyond} assess how well predicted confidence intervals align with observed distributions, though ECE’s binning requires adaptive methods for reliability.

Selecting metrics depends on the task and dataset. A comprehensive evaluation combines prediction accuracy, distributional fit, and uncertainty quantification. Recent work favors WD and MMD over NLL for assessing generative quality.

\section{Applications}
\label{sec:Applications}
The existing literature on neural STPPs mainly underscores their applications in public safety and urban mobility. 
%These applications can focus on either predicting events or delivering actionable insights based on event dynamics.

\paragraph{Natural disasters.} Neural STPPs model earthquake and wildfire occurrences by capturing spatiotemporal dependencies. The ETAS model is widely used for earthquake forecasting, with neural extensions improving prediction. \citet{nicolis2021prediction} enhance ETAS using neural networks for seismic forecasting, while \citet{zhang2024combining} integrate it with deep learning. For wildfires, \citet{xu2023spatio} incorporate environmental factors and remote sensing data to predict ignition probability and magnitude, demonstrating strong performance on California data.

\paragraph{Crime.} STPPs are increasingly used for crime prediction. \citet{dong2024atlanta} model gun violence in Atlanta with a non-stationary Hawkes process, integrating socio-economic covariates to improve predictions. \citet{dong2024spatio} develop a spatiotemporal network point process for Valencia crime, mapping events onto streets to better capture urban dynamics.

\paragraph{Traffic.} \citet{zhu2021spatio} introduce an attention-based STPP integrating traffic sensor data and 911 call records to model congestion dynamics, capturing self-excitation and external influences. They use an NLP-inspired attention mechanism for temporal dependence and a `tail-up' approach for spatial correlations on road networks. \citet{jin2023spatio} propose a congestion prediction model combining GCNs for spatial dependencies, Transformers for temporal patterns, and a continuous GRU with neural flow for instantaneous traffic behavior.

\paragraph{Epidemiology.} STPPs model infectious disease spread by capturing event transmission dynamics. \citet{li2021understanding} propose an intensity-free STPP using generative adversarial imitation learning, while \citet{dong2023non} develop a non-stationary STPP with an NN kernel to model heterogeneous COVID-19 case correlations and spatial variations.

\section{Open Challenges}
\label{sec:Open Challenges}

Neural STPPs have advanced event modeling but still face key obstacles—computational constraints, interpretability, reproducibility, and real-world applicability. Overcoming these will enable robust, scalable, and interpretable models.

\paragraph{Reproducibility.}
A significant barrier to advancing neural STPP research is the lack of standardized experimental setups and consistent baseline comparisons. In contrast to TPPs which benefits from unified libraries like \cite{xue2024easytpp}, there is no comprehensive implementation for spatial and temporal methods in STPPs. A unified library incorporating diverse neural architectures, metrics, and training strategies would facilitate better ablation studies and benchmarking. While some perform well under specific conditions or on curated datasets, fair evaluations across architectures remain challenging without robust testing environments. Additionally, the limited availability of open-source tools further restricts wider adoption and reproducibility of these methods.
% \paragraph{Lack of Standardized Data Sources}
% A major gap in the literature is the absence of standardized benchmarking datasets. Existing datasets often suffer from selection bias, missing data, and varying levels of granularity, complicating integration and analysis. Different datasets exhibit diverse phenomena, such as spatial heterogeneity, long-range dependencies, and entangled spatiotemporal dynamics. Some datasets display self-exciting behavior, while others show self-correcting patterns. The lack of a unified dataset with consistent testing conditions means benchmarking efforts are often limited to specific attributes or datasets where a method performs well. A comprehensive event database would not only standardize benchmarking but also drive methodological advancements and improve event prediction metrics. Additionally, the lack of contextual datasets hinders efforts to develop foundational models for event prediction, which could improve the understanding of event dynamics in data-scarce environments.

\paragraph{Benchmarking.}
% A major gap in the literature is the absence of standardized benchmarking datasets. Existing datasets often suffer from selection bias, missing data, and varying levels of granularity, complicating integration and analysis. Different datasets exhibit diverse phenomena, such as spatial heterogeneity, long-range dependencies, and entangled spatiotemporal dynamics. Some datasets display self-exciting behavior, while others show self-correcting patterns. The lack of a unified dataset with consistent testing conditions means benchmarking efforts are often limited to specific attributes or datasets where a method performs well. A comprehensive event database would not only standardize benchmarking but also drive advancements and improve event prediction metrics. Additionally, the lack of contextual datasets hinders efforts to develop foundational models for event prediction, improving the understanding of event dynamics in data-scarce environments.
A major gap in the literature is the absence of standardized benchmarking datasets. Existing datasets often suffer from selection bias, missing data, and varying granularity, complicating integration and analysis. Different datasets exhibit diverse phenomena, such as spatial heterogeneity, long-range dependencies, and entangled spatiotemporal dynamics. Some datasets display self-exciting behavior, while others show self-correcting patterns. The lack of a unified dataset with consistent testing conditions means benchmarking efforts are often limited to specific attributes or datasets where a method performs well. A comprehensive event database would aid in model development and evaluation. Additionally, the lack of contextual datasets hinders efforts to develop foundational models for event prediction, improving the understanding of event dynamics in data-scarce environments. 
%TODO: Add chronos citation

\paragraph{Architectures.}
Generative strategies and alternative training objectives represent promising yet underexplored avenues for STPP research. While GAN-based architectures and Wasserstein objectives have shown promise in temporal point processes \citep{xiao2017wasserstein}, their adaptation to spatiotemporal applications remains limited. Similarly, methods moving beyond MLE, particularly those leveraging Transformer-based models instead of RNNs, could enhance the flexibility and robustness of event modeling. Continuous-time neural TPPs, such as those proposed by \citep{bilovs2021neural} and \citep{chen2024contiformer}, have demonstrated potential for temporal modeling but have yet to be sufficiently applied to spatial event modeling. Many existing neural STPPs treat space and time independently, missing opportunities to improve temporal predictions by capturing spatial relationships. Integrating GNNs to enhance spatial modeling remains an open challenge.

\paragraph{Applicability.}
Despite improvements in predictive performance, neural STPPs are rarely utilized to inform policy decisions or design interventions. Their lack of interpretability limits their applicability in real-world scenarios, where generalizable, interpretable models with uncertainty quantification are essential. Policymakers often seek to understand the impact of interventions, necessitating models that incorporate contextual factors and causal effects in event propagation. This underscores the need for research in interpretable neural networks, neuro-symbolic methods, and counterfactual analysis for retrospective policy evaluation and deeper insights into event modeling.

\paragraph{Causality and uncertainty.}
Policy-focused applications of neural STPPs require deeper insights into event propagation and the causal factors influencing event rates. However, causal inference and uncertainty quantification remain underexplored in this field. Bayesian methods, which could integrate expert knowledge through priors, have not been widely adopted in neural spatiotemporal event modeling. Similarly, causal representation learning must address spatiotemporal confounding, but current research has yet to fully integrate these methods into neural STPPs \citep{wang2024discovering}. Advancing these areas is critical for developing models that can inform decision-making and intervention design.

\section*{Ethical Statement}

%There are no ethical issues to declare.
% Feel free to revert back to the prior statement
% Point processes are applied in some situations where ethical considerations must be taken into account. 
% For example, point processes have been used to allocate police patrols (\cite{mohler2015randomized}), and such algorithms may exhibit racial biases that could lead to disparate impacts \cite{alikhademi2022review}.  Unintended consequences and harms should be considered before Neural STPPs are used as the basis of spatial interventions.
Point processes are used in ethically sensitive areas, such as police patrol allocation \citep{mohler2015randomized}, where biases may cause disparate harms \citep{alikhademi2022review}. Unintended consequences should be considered before using Neural STPPs for spatial interventions.

%\section*{Acknowledgments}

\newpage
\bibliographystyle{named}
\bibliography{ijcai25}

\end{document}